\def\year{2021}\relax
\documentclass[letterpaper]{article} 
\usepackage{aaai21}  
\usepackage{times}  
\usepackage{helvet} 
\usepackage{courier}  
\usepackage[hyphens]{url}  
\usepackage{graphicx} 
\usepackage{natbib}  
\usepackage{caption} 
\usepackage{amsfonts}
\usepackage{amsmath}
\usepackage{color, soul}
\usepackage{booktabs}
\usepackage{tabularx}
\usepackage{siunitx}
\usepackage{arydshln}

\title{Predictive Control Using Learned State Space Models\\
via Rolling Horizon Evolution}
\author{
    Alvaro Ovalle and Simon M. Lucas\\
}
\affiliations{
    School of Electronic Engineering and Computer Science\\
    Queen Mary University of London \\  

    London, United Kingdom\\
    \{a.ovalle, simon.lucas\}@qmul.ac.uk

}

\begin{document}

\maketitle

\begin{abstract}
    A large part of the interest in model-based reinforcement learning derives from the potential utility to acquire a forward model capable of strategic long term decision making. Assuming that an agent succeeds in learning a useful predictive model, it still requires a mechanism to harness it to generate and select among competing simulated plans. In this paper, we explore this theme combining evolutionary algorithmic planning techniques with models learned via deep learning and variational inference. We demonstrate the approach with an agent that reliably performs online planning in a set of visual navigation tasks.
\end{abstract}

\section{Introduction}


\noindent The capacity to plan confers an agent several advantages such as counterfactual reasoning, the evaluation of different courses of action or the possibility to prepare for future contingencies. In order to plan, a model that captures the relevant dynamics of a task or an environment must be accessible. However, for many complex tasks or unknown situations a model might not be initially available. In these cases it becomes crucial to learn from experience the necessary knowledge that could support the planning mechanisms. 

Recent advances in the modeling of temporal sequences through deep learning have generated a renewed interest in model-based reinforcement learning (MBRL), as a potential path to acquire a model of the environment in its absence, and integrating both aspects of the interaction (i.e. learning and planning) within a broader unified process. 

As the capacity to learn more powerful predictive models increases, the possibility to leverage some of the tools and techniques that have been developed for planning becomes more relevant. Among the recent works in MBRL, some have relied on Monte-Carlo Tree Search (MCTS) \cite{coulom2006} when the interest is in harnessing a learned model operating with discrete action spaces \cite{schrittwieser2020b}, while for continuous action spaces the cross-entropy method (CEM) \cite{rubinstein1999} has been a common choice for updating the distribution used to sample action sequences \cite{chua2018a,hafner2019a}. Alternatively, other works have opted for learning RL policies from the simulated outputs produced by the model \cite{hafner2019b,lee2020c}. 

In this paper we present rolling horizon evolution (RHE) \cite{perez-liebana2013} as a viable alternative to plan and guide decision making in discrete action spaces. In particular, we integrate RHE with state space models (SSM) learned from raw pixel observations. We verify the performance in a set of navigation tasks, where the agent deals with global or local observation spaces and stochastic elements.

\section{Rolling Horizon Evolution}


Rolling Horizon Evolution (RHE) \cite{perez-liebana2013} is a family of general real-time planning algorithms with close connections to Model Predictive Control (MPC). The idea behind RHE is the application of evolutionary algorithmic techniques to action sequences. The process consists in the random generation of $N$ action sequences of length $H$ which then are manipulated by genetic operators (e.g. mutation and crossover). Each of the candidate sequences is executed inside a forward model up to the planning horizon $H$, unless a termination condition is prematurely triggered. The sequences are evaluated according to a given heuristic or score function to obtain their fitness. The evolutionary process of altering, evaluating and selecting the sequences is repeated for a $G$ number of generations. Once it is over, the first action from the highest ranked sequence is executed in the actual environment and the cycle is repeated.

Although RHE was originally devised, and has often been used in conjunction with perfect simulators, it can be adapted to a broader type of situations due to its generality, as long as it is possible to instantiate a forward model to simulate and evaluate the rollouts. Naturally, this also implies that it is largely dependent upon the quality of the model that it has access to. For the situations that we are concerned in this manuscript, when a forward model is not initially available, RHE has been able to interoperate with approximated or imperfect forward models \cite{lucas2019b,ovalle2020a,ovalle2020,olesen2020}

\section{State Space Models in Model-Based Reinforcement Learning}

\begin{figure*}[t]
    \centering
    \includegraphics[width=0.7\textwidth]{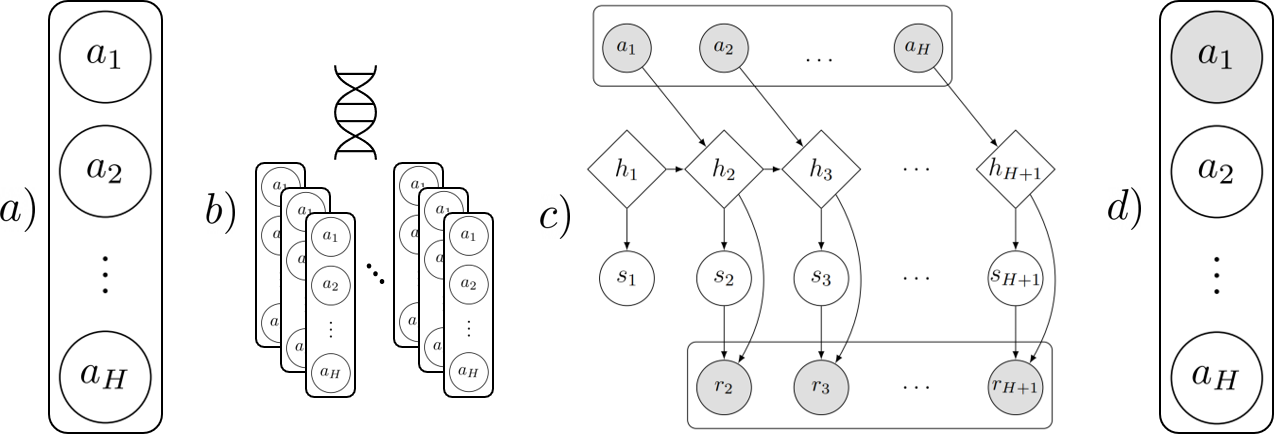} 
    \caption{Integration of RHE (RMHC version) and a SSM. a) Sample an action sequence. b) Apply genetic operators to generate candidates. c) Simulate trajectories in latent space by executing each of the sequences and gather their corresponding $r_{2:H+1}$ to evaluate them. d) From the selected sequence, the agent takes $a_1$ and executes it in the actual environment.}
    \label{rhe_rssm}
\end{figure*}

One of the most challenging aspects of MBRL is that the model that is acquired must be sophisticated enough to capture the dynamics of the environment while being flexible and robust to potential errors. Some research has focused on producing faithful reconstructions \cite{kaiser2020,bamford2020b} while others have opted instead on learning to predict specific aspects of the RL framework \cite{schrittwieser2020b} or the environment \cite{freeman2019} that are thought to be necessary to perform a particular task.

Regardless of the approach, an agent must deal with the uncertainty that arises either from its epistemic shortcomings or from stochastic components attributed directly to the environment. This has motivated the combination of probabilistic modeling with deep reinforcement learning. A particularly active research direction in MBRL has been to use variational inference methods for training state space models (SSM) \cite{buesing2018a,hafner2019a,hafner2019b,okada2020b,lee2020c}. The SSMs \cite{kalman1960} are a class of sequential latent variable models that consider a hidden (i.e. latent) state, here denoted as $s$, which summarizes essential aspects of the past.

Let us then consider an environment that at each time step $t$ generates an output $x_t$ sensed by an agent, containing a visual observation $o_t$ and a reward $r_t$. The agent interacts with the environment by carrying out an action $a_t$. Instead of trying to approximate directly $p(x_{t+1}|x_{\leq t},a_{\leq t})$ the agent can learn a transition model based on $p(s_{t+1}|s_t,a_t)$. From the point of view of planning, this offers the advantage that during a simulated trajectory it is not a requirement to reconstruct a predicted observation $o$ as the rollout can take place in the more compact space representation of $s$, making the process less computationally demanding. In order to infer the latent states, the agent uses its previous observations and actions to learn an encoder $q(s_t|o_{\leq t}, a_{<t})$. In addition, it also learns the emission model which can be decomposed into the observation model $p(o_t|s_t)$ and the reward model $p(r_t|s_t)$.

\subsection{Recurrent State Space Models}

In this paper we focus on a particular kind of SSM characterized in \citep{hafner2019a}, which splits the hidden state into two elements. The stochastic state $s$ that allows the agent to handle model uncertainty by having the capacity to simulate multiple alternative futures, and the deterministic state $h$ which is suggested to assist the capacity of the model to store, access and transport information more robustly. The deterministic component is implemented through a recurrent neural network giving the model its name of recurrent state space model (RSSM). This rearrangement in the structure of the graphical model brings some differences as we now have $h_{t}=f(h_{t-1}, s_{t-1}, a_{t-1})$ and $s_t \sim p(s_t|h_t)$ as dependencies to obtain the observation model $p(o_t|s_t,h_t)$ and the reward model $p(r_t|s_t,h_t)$.

\section{Rolling Horizon within SSMs}

There have been several enhancements, strategies and hybrid formulations proposed in the RHE literature \cite{horn2016,gaina2017b,gaina2017}. With the exception of the \textit{shift-buffer}, which is explained below, we try to maintain the underlying planning mechanics as essential as possible and consider a minimal evolutionary scheme based on the (1+1) random mutation hill climber \cite{lucas2019c}. A single action sequence $(a_t, \dots, a_{t+H})$ is uniformly sampled from the space of actions, while the other $N-1$ sequences are obtained by mutating this first sequence. As previously elaborated, the sequences are executed to simulate a trajectory by instantiating a forward model. To evaluate the sequences we use a reward function that takes the hidden states $s_t,h_t$ and the action $a_t$. Nonetheless in principle it could be substituted by another utility function or by intrinsic signals that, for instance, could direct the agent to regions of the environment where the agent might expect to increase its knowledge and improve the predictive model \cite{ovalle2020}. The evaluation stage can be summarized as,

\begin{equation}
    \pi(s,h) = \underset{a_{t:t+H} \in \mathcal{P}_t}{\arg\max} \mathbb{E} \bigg[r(s_t,h_t,a_t) + \sum_{i=1}^H r(\hat{s}_{t+i},\hat{h}_{t+i},a_{t+i})\bigg]
\end{equation}

Note that the difference between $s_t$ and the subsequent $\hat{s}_{t+1:H}$ is that the former is inferred from current observation using the encoder. Whereas the latter are generated entirely in latent space by querying $p(s_{t+1:H}|h_{t+1:H})$. The agent executes the first action of the highest evaluated sequence available in the current population $\mathcal{P}_t$. For the next action instead of replanning from scratch by discarding the previous plan, RHE applies a \textit{shift-buffer} by shifting the sequence one step to the left and appending a random action at the end. The sequence then is used as the seed and the process is repeated again. For the world model we are interested in learning an approximation to $p(o_{0:T}|a_{0:T})$. Thus following standard variational techniques \cite{jordan1999}, the RSSM is trained to maximize the evidence lower bound (ELBO) of the log likelihood of the model,

\begin{equation}
    \begin{split}
        \ln p(o_{t:B}|a_{t:B}) &\geq \sum_t^B \mathbb{E}_{q(s_{t}|o_{\leq t},a_{<t})} \big[ \ln p(o_{t}|s_{t}) \\
        & - D_{KL}[q(s_t|o_{\leq t},a_{<t})||p(s_t|s_{t-1},a_{t-1})] \big]    
    \end{split}
\end{equation}

\section{Experiments}

\subsection{Environment}

\begin{figure}[t]
    \centering
    \includegraphics[width=0.49\columnwidth]{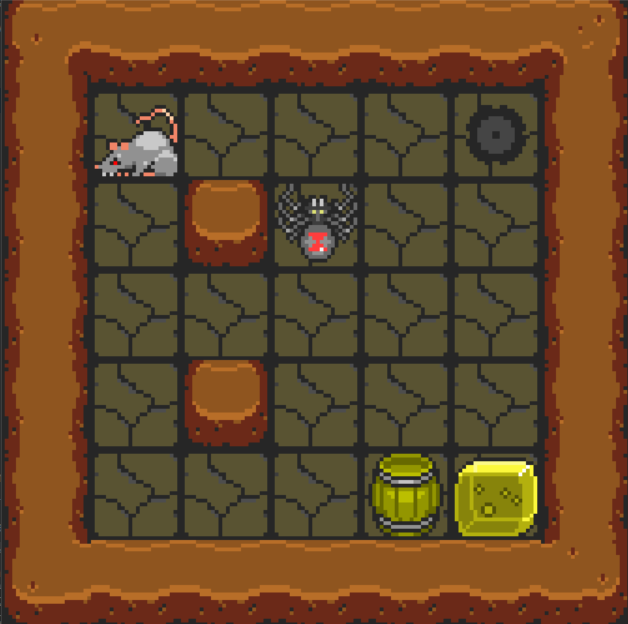}
    \includegraphics[width=0.49\columnwidth]{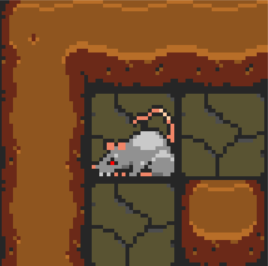} 
    \caption{The environment. The frames show the two type of observations that an agent can receive. On the left showing the whole grid and on the right its local neighborhood.}
    \label{fig:env}
\end{figure}

We verify the ability of the agent to learn a forward model while simultaneously being able to search, select and execute adequate plans in a top-down grid-based visual task designed with Griddly \cite{bamford2020a}. The environment consists of a $7\times7$ grid corresponding to $168\times168$ RGB pixels. The input that the agent receives is a rescaled raw pixel observation of $3 \times 64 \times 64$. The environment is illustrated in the fig. \ref{fig:env} where the agent, represented by the rat, starts the episode in the upper left corner. The task is a classic navigation task where the agent must discover the goal location while avoiding obstacles and possible sources of early termination. The goal here is given by the cheese in the lower right corner, if the agent reaches this position the agent receives a reward of $+1$ and the episode terminates. We conduct four variations of the experiment focusing on two main different aspects of the task. The first of those, is the inclusion of stochastic features in the environment. For this version of the task, we include an antagonistic element represented by the spider that moves randomly throughout the grid. Whenever an encounter between them occurs, the agent receives a reward of $-1$ and the episode terminates. The task also terminates with a reward of $-1$ if the agent reaches the hole located in the upper right corner. For the non-stochastic version of the task the spider is absent and success or failure conditions are only triggered when the agent reaches the goal or the hole positions respectively.

The second aspect we explore is the capacity to deal with global and local observations, and whether the agent can learn a local forward model that can support decision making given the setup. For this variation, the agent visualizes its surrounding $3 \times 3$ cell neighborhood as opposed to the whole grid (fig. \ref{fig:env} right). Accordingly, instead of receiving a raw pixel input of $3 \times 64 \times 64$ it will receive a $3 \times 28 \times 28$ array. All four tasks have a maximum time limit of $500$ steps, if reached, the episode terminates with a reward of $0$ for the agent.

\subsection{Technical Details}

The agent is trained end-to-end and starts by gathering data following a random policy for a few episodes to populate the replay buffer. Then the agent alternates between a learning, and a planning and acting phase. During the learning phase the agent samples data from the buffer to train the RSSM, which entails training the following components:

\begin{itemize}
    \item an inference model $s_t \sim q_{\phi}(s_t|h_t,o_t)$.
    \item the deterministic $h_t=f^{RNN}_\theta(h_{t-1},s_{t-1},a_{t-1})$ and stochastic $s_t \sim p_\theta(s_t|h_t)$ components.  
    \item the observation $o_t \sim p_\lambda(o_t|h_t,s_t)$ and the reward model $r_t \sim p_\xi(r_t|h_t,s_t)$. 
\end{itemize}

The whole neural network architecture is trained to maximize the ELBO via stochastic gradient variational inference \cite{kingma2014a,rezende2014a}. During the planning phase the agent queries the approximate forward model that it currently possesses by observing $o_t$ and inferring $\{s_t,h_t\}$. All the observations gathered during the episode are also stored in the buffer for subsequent learning phases. The appendices contain additional details about the architecture deployed during the experiments.

\subsection{Results}

\begin{figure*}[t]
    \centering
    \includegraphics[width=0.88\textwidth]{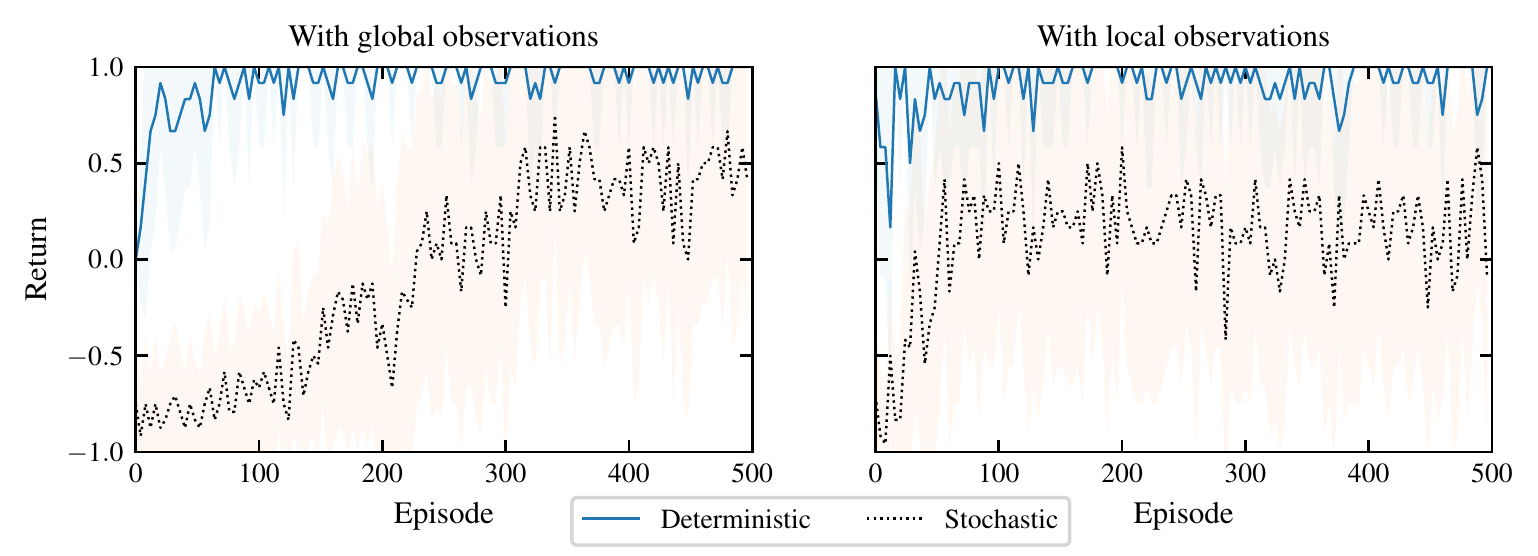}
    \caption{Performance in the task during 500 episodes. The plots show mean and standard deviation over three seeds. The left shows the performance of the agent when it observes the whole grid, while on the right the agent only observes its local neighborhood.}
    \label{fig:res}
\end{figure*}

\subsubsection{Deterministic}

\begin{figure}[htb]
    \centering
    \includegraphics[width=0.8\columnwidth]{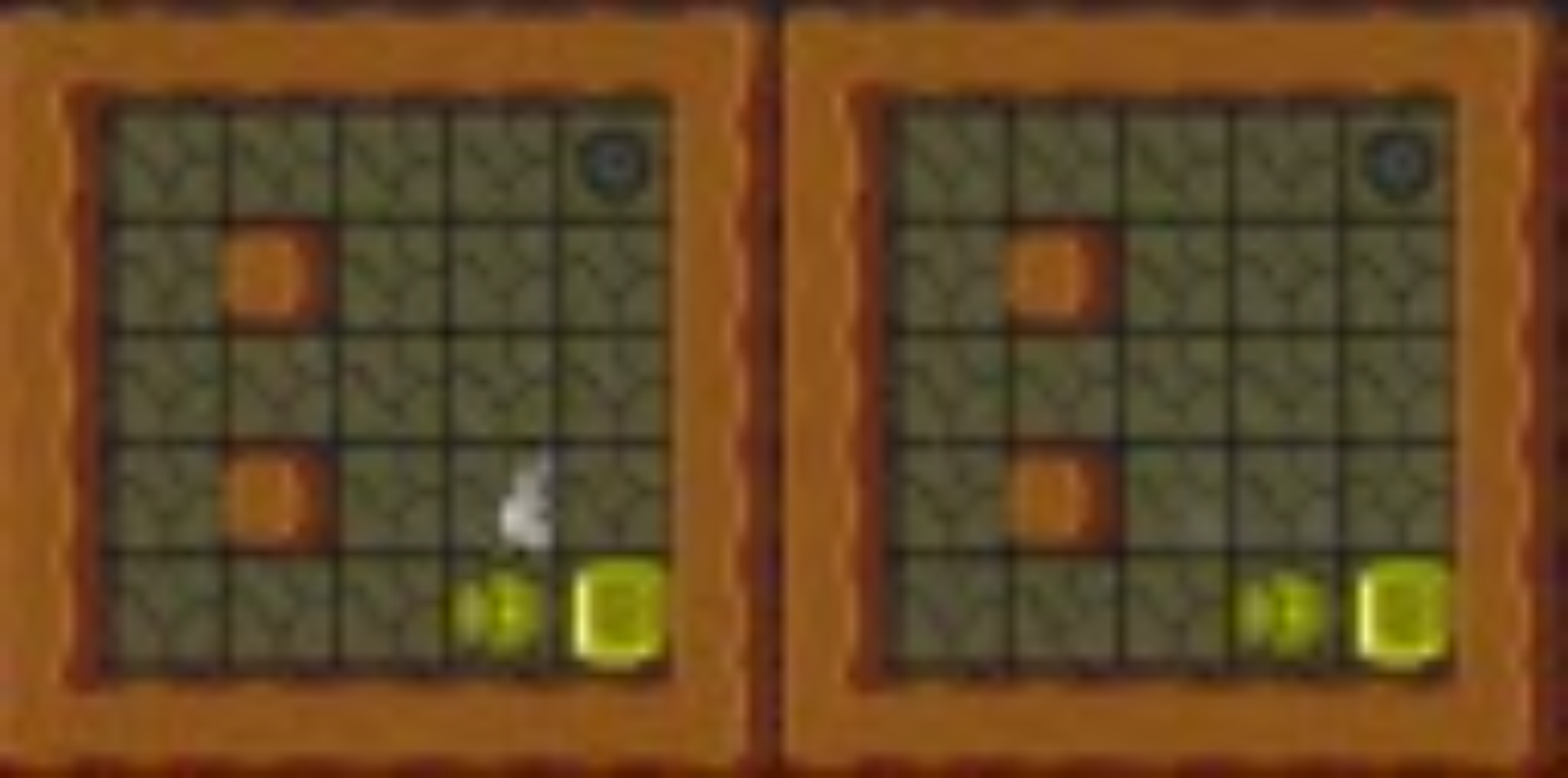}
    \includegraphics[width=0.8\columnwidth]{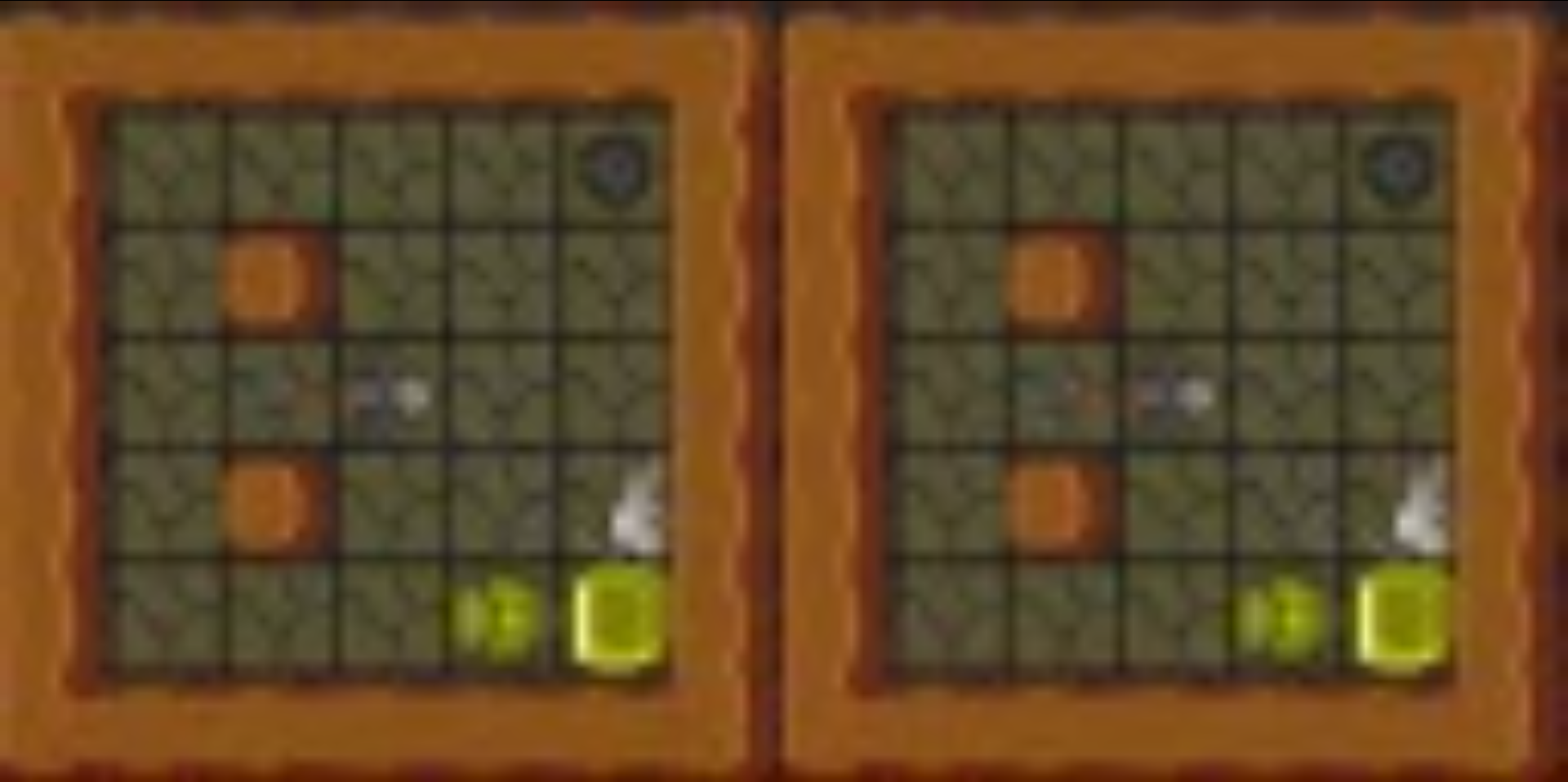}
    \includegraphics[width=0.8\columnwidth]{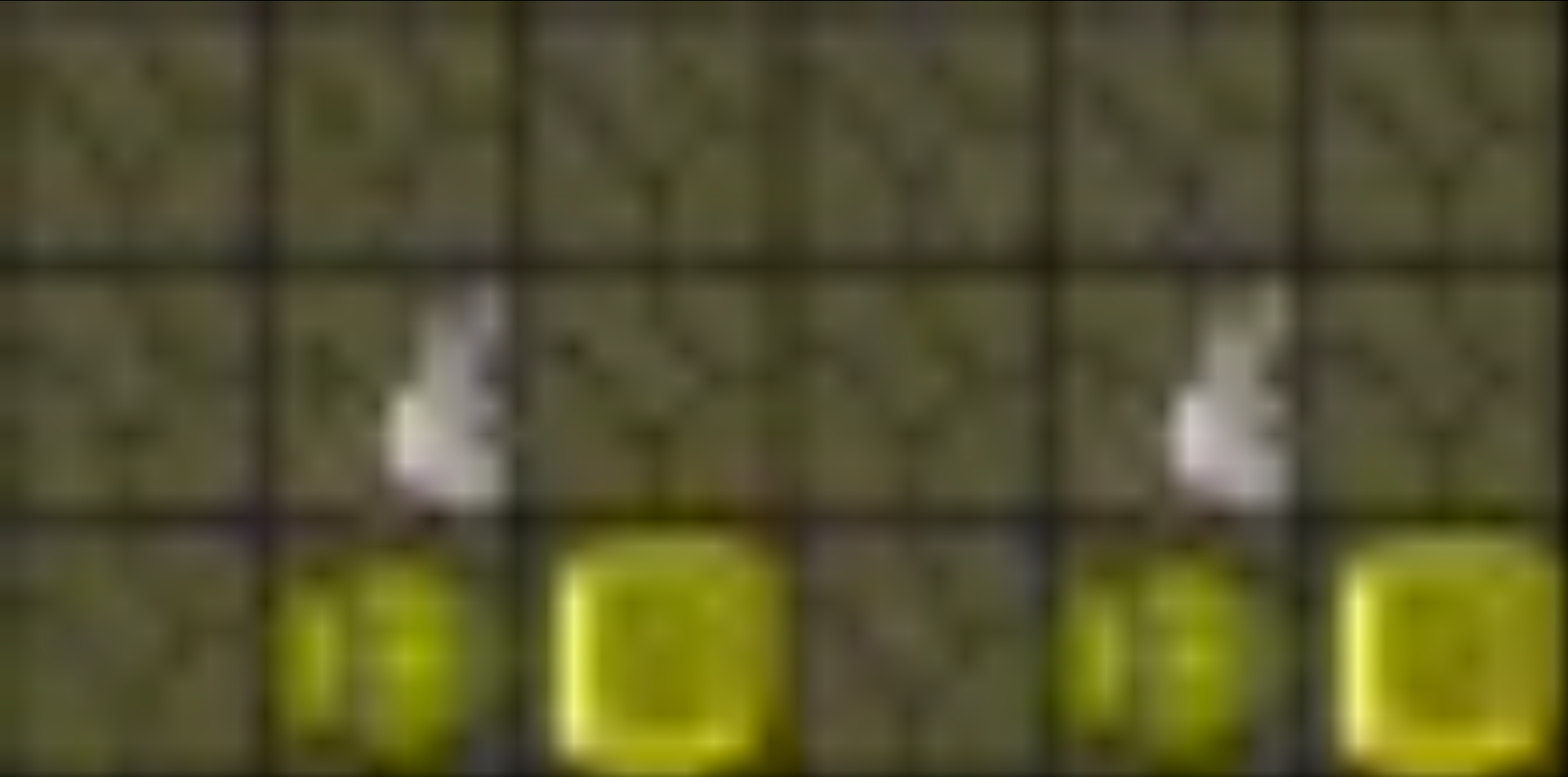} 
    \caption{Examples of actual (left column) and reconstructed (right column) frames. Top: the agent is able to solve the task even before an accurate visual reconstruction can be extracted from the internal model. Middle: example of the agent performing and modelling the task. In this case, the spider serves as an stochastic element in the environment. Bottom: an agent acting in the environment and modelling its local neighborhood.}
    \label{fig:recon}
\end{figure}

For the deterministic version of the environment both agents have comparable task performance independently from whether the observations they receive are global or only restricted to its immediate neighborhood. Both agents achieve proficiency early during training, however the agent that receives local observations achieves it more efficiently as it uses less steps. Although later during training, both agents use an equivalent number of steps to reach the goal. During the analysis of the behavior of the agent, it was observed that it achieved competence in the task well in advance of learning how to reconstruct accurately its environment. This suggests that the latent representations learned by the agent prioritize the encoding of features relevant for reaching (or avoiding) favorable (or undesirable) states and this might not necessarily have a perceptible graphical translation (fig. \ref{fig:recon} top). That is, the usefulness of a model should not be dictated by its capacity to replicate the environment with meticulous precision.

\subsubsection{Stochastic}

When the environment included an stochastic element, the agent that only receives local observations again exhibited higher initial responsiveness. However on the long term it is the agent with a more global forward model that is able to carry out plans with a higher chance of succeeding. Analyzing the behavior of the agent and the model reconstruction offered potential insights and interpretations. The agent that receives global observations seems to develop an initial aversion towards moving to the rest of the grid, instead preferring to move only vertically to remain close to the starting position and behind the blocks. During these earlier stages the plans executed by the agent seem exclusively intended to avoid encounters with the enemy. Since this element is a defining factor into whether the agents succeeds or not in the task, it is perhaps until the model attempts to capture more of the dynamics of the agent with respect to the spider that it starts executing plans to move towards other positions in the grid and eventually towards the goal. In contrast, the agent that operates with local observations does not have a continuous necessity to model the location or dynamics of the spider, as most of its observations do not contain it and thus favors less conservative plans. As previously mentioned, later during training it is the agent operating with a global model the one that attains the goal more reliably, as it has learned to consider more complex dynamics used to avoid the spider. Overall these initial assessments motivate a further elucidation of the relation between relevance and representations, observations and objectives.

\section{Discussion}

The design of architectures based on SSMs has grown in interest within MBRL due to the temporal structure and the consideration of a latent space that are intrinsic to the formalism. The latter is not only important for filtering out non-essential elements of the interaction between the agent with the environment but also for the acquisition of economical and compact representations that can facilitate a more expedite decision making. In this paper we presented how to harness these representations to support real-time planning in discrete action tasks via RHE. Although these planning techniques have largely been developed for tasks where the access to a perfect simulator is assumed, the principles are general enough that can be extended as long as the signals predicted by the model are sufficiently reliable to allow RHE to evaluate competing plans. 

An aspect that facilitated the integration is the modularity of both components. In RHE the generation of the action sequences occurs independently from their evaluation, and in the RSSM learning the transition model remains partially separated from learning the reward model. This implies that after the sequences of latent states have been gathered they can be evaluated concurrently. This is specially significant for search-based algorithms that must balance time and computational constraints with the capacity to evaluate a large number of potential plans. Another affordance granted by this modularity is that the sequences of latent states could be reused by RHE to invoke on demand other components in the architecture that could predict additional signals to assist during the evaluation. 

From the point of view of the architecture, compared with muzero \cite{schrittwieser2020b} or \textit{imagination-based} RL \cite{hafner2019b, lee2020c}, the approach presented here only requires training the SSM and not additional components such as a policy or a value network. This could be preferred from a computational perspective depending on the circumstances. Nonetheless there are no structural obstacles to incorporate additional networks or to use learned value functions within the evaluation criteria of RHE, and indeed it could be particularly beneficial for more complex tasks. Exploring these ideas and establishing these comparisons more formally could be expanded in future work.

\section{Acknowledgments}

We thank Ercüment İlhan for helpful discussions and to Chris Bamford for his assistance with Griddly. This research utilised Queen Mary's Apocrita HPC facility, supported by QMUL Research-IT. http://doi.org/10.5281/zenodo.438045

\section{Appendix}

\subsection{Reconstructions}

\subsection{Architecture}\label{apx:arch}

\subsubsection{Observation embedding} A $3 \times 64 \times 64$ or a $3 \times 28 \times 28$ array, for the global and local case respectively, is passed through a four-layer convolutional network with output sizes of $32,64,128$ and $256$, kernels of $4 \times 4$ and stride $2$. The exception is when the network receives a local observation, if that occurs, the kernel in the last convolutional layer is of size $1$. All the activation functions are ReLU. The output is passed further to a linear layer to produce an embedding of size $200$.

\subsubsection{Transition model} An embedding of the stochastic state, action and deterministic state is passed through a GRU to obtain the next deterministic state. The next stochastic state is sampled by applying the reparameterization trick. 

\subsubsection{Observation model} For decoding an observation an embedding of the latent states is passed through a deconvolutional network with four layers with outputs $128, 64, 32$ and $3$, kernels $5,5,6$ and $6$, and stride $2$ for global observations. For local observations the kernels are $2,3,5$ and $4$. ReLU is used for all activation functions.

\subsubsection{Reward model} The embedding of the latent states is passed through a three-layer feed forward neural network with hidden size $200$ and ReLU activation functions.

\subsection{Training}\label{apx:train}

The architecture is trained end-to-end combining a testing stage, in which the agent acts in the environment for one episode, and a training stage. For the latter, the SSM architecture is trained for $100$ epochs from data sampled from the buffer that contains the observations that have been collected by the agent. The architecture is trained to maximize the ELBO via stochastic gradient descent using the Adam optimizer.

\subsection{Hyperparameters}\label{apx:hyp}

\begin{center}
\begin{tabularx}{\columnwidth}{Xr} \toprule
    \textbf{Training} &  \\ \midrule
    \textbf{Learning rate}  &  6e-4\\ \hdashline[0.5pt/5pt]
    \textbf{Epsilon}  &  1e-4\\ \hdashline[0.5pt/5pt]
    \textbf{Grad clip norm}  &  1000\\ \hdashline[0.5pt/5pt]
    \textbf{KL weight}  &  0.1 \\ \hdashline[0.5pt/5pt]
    \textbf{Free nats}  &  3\\ \hdashline[0.5pt/5pt]
    \textbf{Epochs}  & 100 \\ \midrule
    \textbf{Embeddings}  &  \\ \midrule
    \textbf{Observation}  & 200 \\ \hdashline[0.5pt/5pt]
    \textbf{Deterministic}  & 200 \\ \hdashline[0.5pt/5pt]
    \textbf{Stochastic}  & 32 \\ \midrule
    \textbf{Replay buffer}  &  \\ \midrule
    \textbf{Buffer size}  & 1e6 \\ \hdashline[0.5pt/5pt]
    \textbf{Seed episodes}  & 5 \\ \hdashline[0.5pt/5pt]
    \textbf{Batch size}  & 50 \\ \hdashline[0.5pt/5pt]
    \textbf{Batch sequence size}  & 20 \\ \midrule
    \textbf{Planning}  &       \\ \midrule
    \textbf{Planning horizon} & 20  \\ \hdashline[0.5pt/5pt]
    \textbf{Candidates} & 300 \\ \hdashline[0.5pt/5pt]
    \textbf{Mutation rate} & 0.5 \\\hdashline[0.5pt/5pt]
    \textbf{Shift-buffer} & True \\ \bottomrule
\end{tabularx}
\end{center}

\bibliography{icaps}

\end{document}